# ESDF: Ensemble Selection using Diversity and Frequency

Shouvick Mondal and Arko Banerjee

*Abstract*— Recently ensemble selection for consensus clustering has emerged as a research problem in Machine Intelligence. Normally consensus clustering algorithms take into account the entire ensemble of clustering, where there is a tendency of generating a very large size ensemble before computing its consensus. One can avoid considering the entire ensemble and can judiciously select few partitions in the ensemble without compromising on the quality of the consensus. This may result in an efficient consensus computation technique and may save unnecessary computational overheads. The ensemble selection problem addresses this issue of consensus clustering. In this paper, we propose an efficient method of ensemble selection for a large ensemble. We prioritize the partitions in the ensemble based on diversity and frequency. Our method selects top K of the partitions in order of priority, where K is decided by the user. We observe that considering jointly the diversity and frequency helps in identifying few representative partitions whose consensus is qualitatively better than the consensus of the entire ensemble. Experimental analysis on a large number of datasets shows our method gives better results than earlier ensemble selection methods.

## I. INTRODUCTION

THE problem of consensus clustering is concerned with combining an ensemble of clusterings to get a qualitatively improved clustering. The need for consensus clustering arises due to the fact that none of the existing clustering algorithms yields satisfactory partition for all instances of input data. Some of these algorithms also give different clustering based on different values of initial parameters and there is no way to determine the most appropriate values of these parameters for a given situation. In such situations, consensus clustering technique attempts to combine the results of different runs to get a better clustering. Most of the consensuses clustering algorithms attempt to heuristically compute a median of all constituents of a given ensemble of clusterings.

While combining multiple partitions, the general interest is to obtain a consensus partition with better quality than that of input partitions. One is not very sure of the size of ensemble that can generate a good consensus and most often the clusterings that constitute the ensemble may be overly generated. Thus it may not be necessary to take all the clustering into account while generating a consensus. The question, then, that naturally arises in this situation is whether we selectively use some of the partitions in the ensemble. In this paper we address this problem. The objective is to devise a method to select a subset of clusterings from a given ensemble so that the consensus on this subset yields as good a clustering as that on the full set in the ensemble. Our method is based on the following metaphor. We view each clustering as the opinion of one expert expressing his opinion of an element's chance being in a cluster. Importance of an opinion is a function of number of experts agreeing on it. We assume that the opinion of major/bigger group is more trustworthy than that of minor/smaller group. On the other hand, if an expert is quite far from agreement with many other experts, his/her opinion may contain some unique characteristics that cannot be ignored for a consensus. Thus a major group with diverse opinion with respect to all other groups is considered to be more valuable information for consensus than that of a major group whose opinion resembles with others. We show here that by prioritizing clusterings on frequency of occurrence and diversity, and selecting the clustering in a greedy fashion in the decreasing order of priority gives a better consensus clustering using substantially less number of clusterings from the ensemble. Experimental investigation reveals that our method achieves statistically significant performance improvements over the entire ensemble for many artificial and real world data sets.

The remainder of the paper is organized as follows. In Section 2, we will review the related literature. Section 3 presents the basic selection strategies based on jointly considering frequency and diversity and their performances are evaluated in Section 4. Finally, we summarize our contributions and conclude the paper in Section 5.

## II. RELATED WORK

The goal of consensus clustering is to combine multiple, diverse and independent clustering arrangements to obtain a single, comprehensive clustering. Formally the problem of combining multiple clusterings can be described as follows: Let S be the set of data points. $S = \{s_1, s_2, …, s_n\}$. We are given a set of $T$ partitions, $P = \{P_1, P_2, …, P_T\}$ of the data points in S. A partition P on S is defined as $P = \{C_1, C_2, …, C_k\}$ such that $C_i \subseteq S$, $C_i \cap C_j = \varnothing$ and $\cup C_i = S$. Our goal is to find a final clustering $P^* = \{C^*_1, C^*_2, …, C^*_k\}$ that optimizes a consensus function. A consensus function maps a given set of partitions $P = \{P_1, P_2, …, P_T\}$ to a final partition P* which is a sort of median of $P_1, P_2, …$ and $P_T$. There are several consensus functions proposed in literature.

Fred *et al* [4] combine clusterings produced by multiple runs of the k-means algorithm into a coassociation matrix. A hierarchical single-link algorithm is used to partition this

Shouvick Mondal, Student ,Dept. of CSE, College of Engineering and Management, Kolaghat, India, Email: shouvick.mondal.cemk@gmail.com
Arko Banerjee,Assistant Professor , Dept. of BSH, College of Engineering and Management, Kolaghat, India, Email: arko.banerjee@gmail.com

matrix into the final consensus clusters. Topchy *et al* [13] formulate the consensus clustering problem into a maximum likelihood problem which is solved by the EM algorithm. Caruana *et al* [2] discuss ensemble selection from a library of trained models. Gionis *et al* [6] provide a formal definition to the problem of cluster aggregation and discuss a few consensus algorithms with theoretical guarantee on quality of the consensus. Topchy *et al* [14] present two approaches using plurality voting and using a metric on the space of partitions. Strehl and Ghosh [12] define the cluster ensemble problem as an optimization problem and maximize Normalized Mutual Information (NMI) of the consensus clustering. They introduce three different algorithms to obtain good consensus clustering, namely Cluster-based Similarity Partitioning (CSPA), HyperGraph Partitioning (HGPA), and Meta-Clustering (MCLA) algorithms. Banerjee and Pujari [17] propose a greedy strategy to select the clusterings in an iterative consensus generation technique that ensures the quality of clustering to be monotonically non-decreasing.

Almost all the algorithms in foregoing discussion attempt to determine a clustering which is in essence a median point in the space of clusterings. This is accomplished by minimizing a disagreement criterion function (or, equivalently maximizing an agreement function). The consensus clustering algorithms differ among themselves in the ways they define the criterion function and the heuristic to optimize the criterion function. However, all these methods take into account the entire set of clusterings to arrive at a consensus clustering. In co-association based methods, the matrices obtained from individual clusterings are aggregated. In hypergraph approach, the entire set of clusterings is used to build a hypergraph and consensus clustering is obtained by cutting the redundant set of hyperedges. In information theory based method, the diversity between a pair of clusterings is defined in terms of Normalized Mutual Information (NMI) and consensus clustering is one that minimizes its total diversity with respect to all input clusterings.

Most often the ensemble is overly generated and computing consensus of a large number of partitions is needlessly time-consuming. In an overly generated ensemble there may be many redundancies and hence judiciously selecting a subset of partitions for consensus computation may result in fast and efficient way of consensus clustering. Fern *et al* [3] address this problem as *Ensemble Selection Problem*. They propose three ensemble selection approaches based on quality and diversity of partitions. The first method, referred to as *Joint Criterion*, proposes a joint objective function that combines both quality and diversity. The second method, *Cluster and Select* (CAS), organizes different solutions into groups such that similar solutions are grouped together. It selects one quality solution as representative of each group and computes the consensus of the representatives. The last method, *Convex Hull*, creates a scatter plot of points, where each point corresponds to a pair of clustering solutions represented by their average quality and diversity, and then uses the convex hull of all points to select solutions. It is shown empirically that among the three methods, *Cluster and Select* method achieves the best overall performance.

## III. SELECTION BASED ON DIVERSITY AND FREQUENCY

In this section we propose a new method of ensemble selection technique which uses the concepts of diversity and frequency of partitions in an ensemble. Let P be the entire ensemble. We assume that each partition in P is generated by one distinct clustering run but P may contain multiple copies of a partition that is generated by multiple distinct runs. Let E be the set that contains all distinct partitions of P. Let us assume E= {$P_1, P_2, ..., P_r$}.

We introduce the following definitions

### A. Definition 1 (Frequency)

The *frequency* of $P_i$ (denoted as $v_i$) in P is the number of occurrences of $P_i$ in P.

### B. Similarity Measures

There have been several proposals to compare a pair of partitions. These include Rand Index [8], Jaccard Index [1], Adjusted Rand Index [7], [11], Wallace Index [15] and Normalized Mutual Information [5], [12]. In this paper we use Adjusted Rand Index as the similarity measure between two partitions. We define this measure below.

Adjusted Rand Index, an important variant of Rand Index, corrects the lack of invariance when partitions are selected at random [7]. Let us first define the following quantities.

$$t_0 = \sum_{r,s} \binom{N_{rs}}{2}, \quad t_1 = \sum_{r=1}^{k_i} \binom{n_{r(i)}}{2}, \quad t_2 = \sum_{s=1}^{k_j} \binom{n'_{s(j)}}{2}, \quad t_3 = \frac{t_1 \times t_2}{\binom{n}{2}}$$

where $N_{rs}$ is the number of common data items in $r^{th}$ cluster of partition $P_i$ and $s^{th}$ cluster of partition $P_j$, that is in $C_r \cap C'_s$. Let $n_{r(i)}$ and $n'_{s(j)}$ be number of items in $C_r$ of $P_i$ and $C'_s$ of $P_j$, respectively.

The Adjusted Rand index is defined as [11]

$$AR(P_i, P_j) = \frac{t_0 - t_3}{\frac{1}{2}(t_1 + t_2) - t_3}$$

For two identical clusterings, the adjusted rand index is 1 and if two clusterings are in total disagreement then the value is 0. We choose adjusted rand index as a similarity measure because it is considered as a good measure of similarity, it is used in many previous clustering studies and it is easy to compute. Note that the selection method we develop in this paper is independent of any particular similarity measure.

### C. Definition 2 (Mean Adjusted Rand Index)

The *mean adjusted rand index* (*MAR*) of a partition $P_i$ with respect to E is defined as follows.

$$MAR(P_i, E) = \frac{1}{(r-1)} \sum_{j=1, P_i \neq P_j}^{r} AR(P_i, P_j)$$

*MAR($P_i$, E)* is an estimate of average similarity of $P_i$ with respect to other partitions in E.

### D. Definition 3 (Diversity measure)

The diversity measure of $P_i$ in E is defined as follows.

$$Div(P_i, E) = 1 - MAR(P_i, E)$$

It is easy to see that if $P_i$ is an isolated partition in E, then $Div(P_i, E)$ is very large(close to 1). On the other hand, if $P_i$ is close to the median of E, then $Div(P_i, E)$ takes on the least value.

### E. Definition 4 (Weight of a partition $P_i$ in E)

For a given clustering $P_i$ in an ensemble E the weight of $P_i$ is defined as follows.

$$w_i = Div(P_i, E) \times \frac{v_i}{\sum_j v_j}$$

$w_i$'s play an important role in certain consensus functions. As discussed earlier, there are several consensus functions most of which are dependent on the underlying diversity measure in the sense that the consensus clustering is a partition that maximizes the aggregated similarity over all partitions in the ensemble. $w_i$ takes into account both diversity and frequency and this in terms of our experts-metaphor, an opinion that is expressed by a number of experts and different from a number of expterts has a higher importance.

## IV. THE ENSEMBLE SELECTION ALGORITHM

The proposed method of ensemble selection starts with E, the set of distinct partitions of P. These distinct partitions are ordered in the decreasing order of their weights $w_i$ and first k partitions are selected for computation of consensus clustering by any of the known methods such as co-association based methods, evidence accumulation based methods, hypergraph based methods, voting based methods. We have experimented with two consensus finding methods, namely CSPA [12] and HGPA [12]. CSPA builds a similarity matrix (also known as co-association matrix) based on the ensemble. In HGPA, the cluster ensemble problem is posed as a partitioning problem of a suitably defined hypergraph where hyperedges represent clusters and the problem is solved by cutting a minimal number of hyperedges.

The pseudo code of our algorithm is given below.

---

INPUT:
    Clustering Ensemble P,
    parameter k
OUTPUT :
    Consensus Clustering
METHOD:
1. Determine E, the set of distinct
    Partitions in P and corresponding frequencies, $v_i$ s.
2. Compute pairwise Adjusted Rand
    Index for each pair of partitions in E
3. Compute *MAR*, mean adjusted
    rand index for each partition
4. For each $P_i$ compute $w_i$
5. Sort the elements of E in
    decreasing order of $w_i$
6. Consider top k partitions in E as E'
7. Compute consensus on E'
    using CSPA/HGPA

---

Pseudo code for ESDF

It is suggested that, if a clustering has high weight then it is more relevant in consensus formation whereas clustering solutions with small weight values, can be considered as outliers of the ensemble and may be omitted. Our motivation of ensemble selection procedure is to generate an ensemble that is high in diversity, less in redundancy and free from outliers. For the remainder of this paper, we refer to our method as Ensemble Selection using Diversity and Frequency (ESDF).

## V. EXAMPLE

As an illustration of the working of ESDF, we consider Iris dataset. For visualization in 2-dimension, the four dimensional data is projected into 2 dimension by Sammon Mapping [10]. Figures 1a, 1b and 1c depict the original clustering, consensus clustering on full ensemble of 100 partitions using CSPA, consensus clustering using ESDF for selection and CSPA for consensus computation, respectively. The Adjusted Rand Index for consensus of whole ensemble and ensemble after selection by ESDF are 0.7140 and 0.8015, respectively. We see that ESDF yields better result compared to considering the entire ensemble of Iris data.

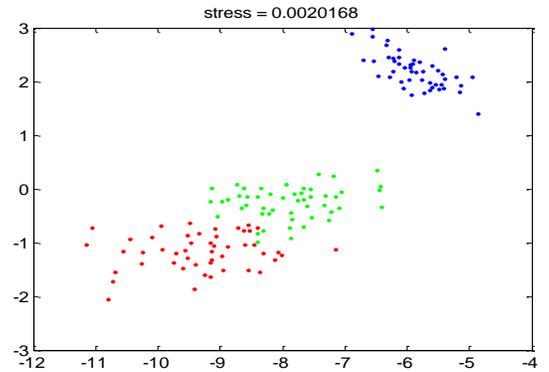

(a)

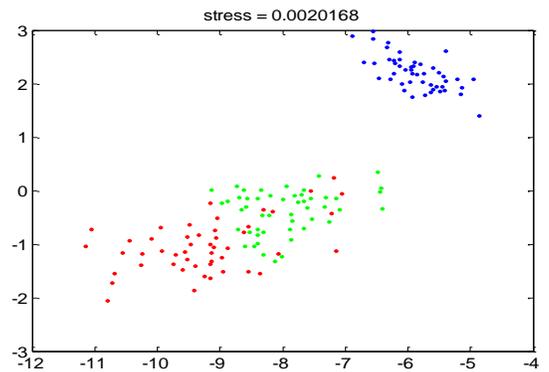

(b)

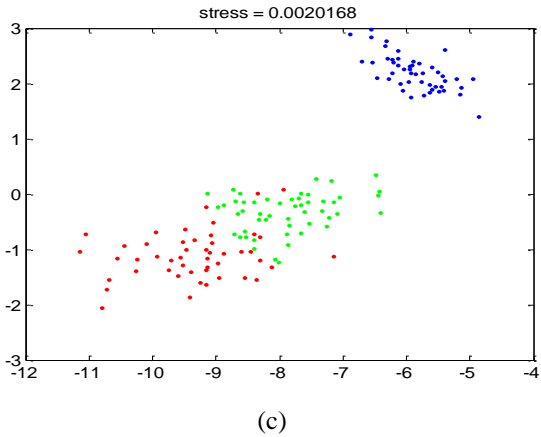

(c)

Figure 1: a. Iris dataset projected in two dimensions. Three different colours represent three ground truth clusters. b. Consensus clustering due to CSPA on entire ensemble. c. Consensus clustering due to CSPA on a subset selected by ESDF procedure.

In this section, we show the importance of joining diversity and frequency together instead of considering these factors individually. We consider three cases.

case1: ensemble selection method using diversity alone.

case2: ensemble selection method using frequency alone.

case3: ensemble selection method by jointly considering diversity and frequency (ESDF).

We show this aspect for Yeast dataset in figure 2. An ensemble of size 100 is generated and partitions are selected from the ensemble in order of (i) diversity, (ii) frequency and (iii) diversity-frequency weight ($w_i$), respectively in three different processes. The Adjusted Rand indices of the output consensus clustering in each of the cases are plotted by varying the size of the ensemble. Figure 2 depicts this result, where the x-axis represents the size of the ensemble and y-axis for Adjusted Rand index. x=c represents the ensemble chosen by removing last 'c' partitions.

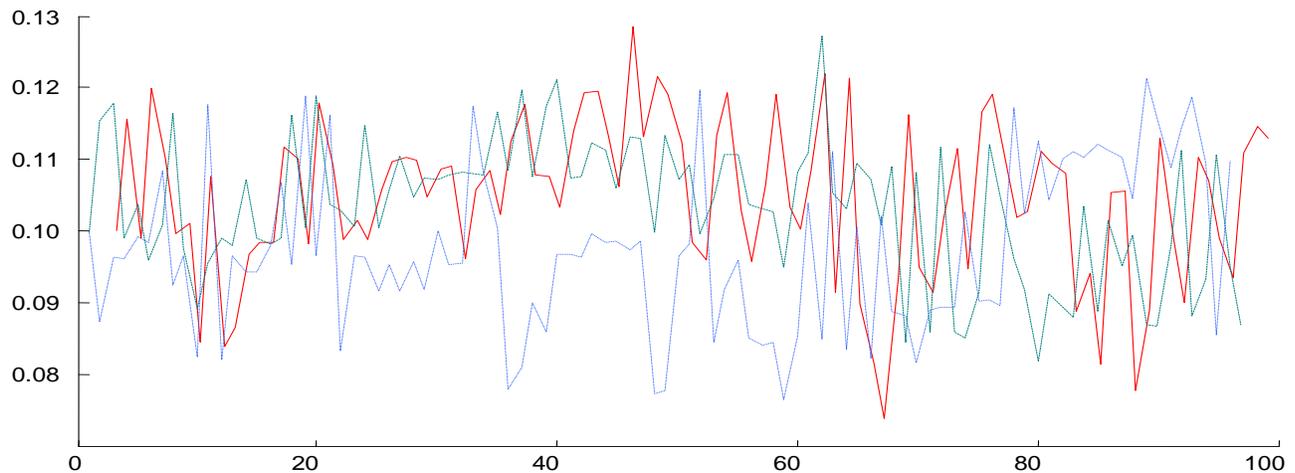

Figure 2: Yeast dataset: ensemble size vs. AR. Ensemble selection using diversity, frequency and diversity-frequency weight are represented by blue, green and red lines, respectively.

The maximum AR values for ensemble selection methods using diversity, frequency and diversity-frequency weight are found to be 0.1273, 0.1213 and 0.1283, respectively. It is vivid from Figure 2, that diversity-frequency weight not only gives the maximum AR value but also the **average AR is higher** than that of other two cases. The same characteristics is exhibited by Chart, Segmentation and Ecoli datasets [16].

## VI. VISUALIZATION USING MANIFOLD LEARNING

To provide an intuitive feel of ESDF algorithm, we use a manifold learning based visualization method. For a pair $P_i, P_j$ of partitions, $1\text{-}AR(P_i, P_j)$ are taken as pair wise distance matrix. Locally Linear Embedding (LLE) [9] on this matrix is computed for the target dimension 5. The LLE is a nonlinear dimensionality reduction method that can be employed for any type of data if the pairwise distances of the objects are known. It preserves the local neighbourhood properties and projects the high dimensional data into lower dimensional space by preserving the coefficients of affine combination of the k-nearest neighbours of each object. In our experiment, we take k=10. Figure 3 and Figure 4 show the embedding of the partitions in ensemble of Ecoli and Chart data in 2-dimensional space (selected 3rd and 4th dimension), respectively. The partitions selected by ESDF and CAS [3] are highlighted with different colours. In the Figures, green dots represent distinct partitions. Red circles represent partitions selected by CAS. Black circles represent partitions selected by our selection algorithm. The red circles with cyan dot represent partitions selected by both CAS and our algorithm. The black square represents ground truth partition. The magenta dot represents the partition due to CSPA on entire ensemble. The black pentagon represents the partition due to CSPA on the ensemble selected by CAS. The blue star represents the partition due to CSPA on the ensemble selected by our algorithm.

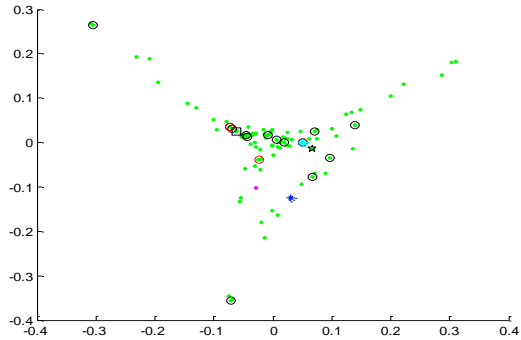

Figure 3: Embedding of the partitions in ensemble of Ecoli data in 2-dimension.

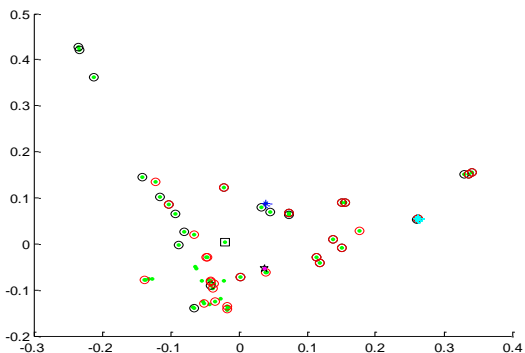

Figure 4: Embedding of the partitions in ensemble of Chart data in 2-dimension.

The LLE based visualization also provides a nice interpretation of CAS. CAS picks up partitions that are representative of groups. One can see from the figures that wherever there is a dense conglomeration of partitions, CAS identifies one of them as representatives. In the same manner, wherever there is a partition that is isolated and frequent, ESDF identifies these partitions for consensus computation.

## VII. EXPERIMENTAL RESULTS

In this section, we examine the performance of the ensembles produced by ESDF and compare them with the performances with cluster and select (CAS) method [3]. First we describe the data sets and the basic settings of our experiments that we use in the evaluation.

### i. Data Sets

Our experiments use real-world data sets CHART, SEGMENTATION, WINE, GLASS, IRIS, YEAST and ECOLI, which are benchmark data sets collected from the UCI machine learning data repository [16]. It should be noted that all the datasets are labeled and contain supervised class information.

Table 1: Basic information of the real world data sets

| Datasets | # Instances | # Features | # Classes |
|---|---|---|---|
| Chart | 600 | 60 | 6 |
| Segmentation | 2100 | 19 | 7 |
| Ecoli | 336 | 7 | 8 |
| Yeast | 1484 | 9 | 10 |
| Iris | 150 | 4 | 3 |
| Glass | 214 | 10 | 6 |
| Wine | 178 | 13 | 3 |
| Vehicle | 846 | 18 | 4 |

The reason that we have chosen those datasets because some are very well known datasets in cluster analysis studies and some are not so well known but due to their high dimensionality they present significant challenge to standard clustering algorithms

### ii. Ensemble Generation

To build our ensemble, we used the k-means algorithm as our base algorithm. K-means is chosen because it is one of the most widely used clustering algorithms and has been used in many previous cluster ensemble studies. Different clustering solutions are obtained by applying k-means to the same data with different random initializations. We generate three different ensembles of size 200 for all datasets. We perform six different tests – three using CSPA and three using HGPA. In the first test CSPA/HGPA is applied on the entire ensemble to obtain a consensus clustering solution. In the second test CAS is applied on the ensemble and then CSPA/HGPA is applied on the resulting ensemble to obtain the final consensus. In the third test we apply ESDF on the ensemble and then CSPA/HGPA is applied on the resulting ensemble to obtain the final consensus solution. To evaluate the experimental performance of we use the pre-existing true class labels and measure the adjusted rand index (AR) between the consensus cluster labels and the true class labels.

In the following tables 'CSPA/HGPA on all' refers to consensus due to CSPA/HGPA on entire ensemble; 'CAS+CSPA/HGPA' refers to consensus due to CSPA/HGPA on a subset selected by CAS; 'ESDF+CSPA/HGPA' refers to consensus due to CSPA/HGPA on a subset selected by ESDF.

The following tables show that ESDF outperforms CAS in case of Chart, Segmentation, Ecoli, Yeast and Iris datasets. Whereas, CAS gives better result in case of Vehicle dataset and almost same results in Glass and Wine datasets (Table not shown). Note, the AR value considered for a particular experiment is the maximum value out of all AR values for each ensemble size.

Table 2: Results for Chart and Segmentation datasets

|  | Chart | | | | Segmentation | | |
|---|---|---|---|---|---|---|---|
|  | Ensemble1 | Ensemble2 | Ensemble3 | | Ensemble1 | Ensemble2 | Ensemble3 |
| CSPA on all | 0.6393 | 0.646 | 0.6521 | | 0.3606 | 0.3552 | 0.478 |
| CAS+ CSPA | 0.7417 | 0.6731 | 0.6890 | | 0.4908 | 0.3798 | 0.4806 |
| ESDF+ CSPA | 0.7816 | 0.7211 | 0.7028 | | 0.5170 | 0.4062 | 0.487 |
| HGPA on all | 0.4573 | 0.4725 | 0.4820 | | 0.1331 | 0.2480 | 0.2395 |
| CAS+ HGPA | 0.5890 | 0.5510 | 0.5532 | | 0.3457 | 0.3888 | 0.3502 |
| ESDF+ HGPA | 0.6028 | 0.5880 | 0.5590 | | 0.3566 | 0.3990 | 0.3562 |

Table 2 shows that on Chart and Segmentation datasets both CSPA and HGPA produce better consensus when ESDF is used as the ensemble selection procedure rather than CAS.

Figure 5 compares ESDF and CAS on Chart dataset with respect to the ground-truth partition. The sixty dimensional dataset is projected into two dimension using sammon mapping. The stress function mentioned at the top of the figures denotes the accuracy of projection.

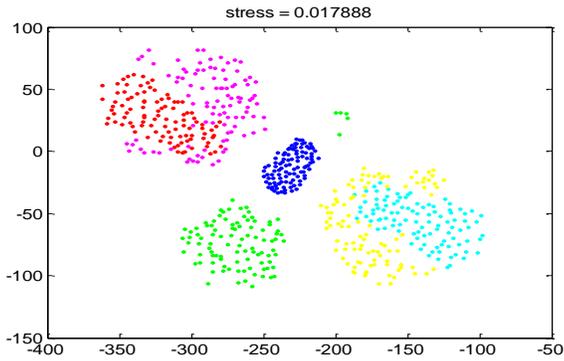

(a)

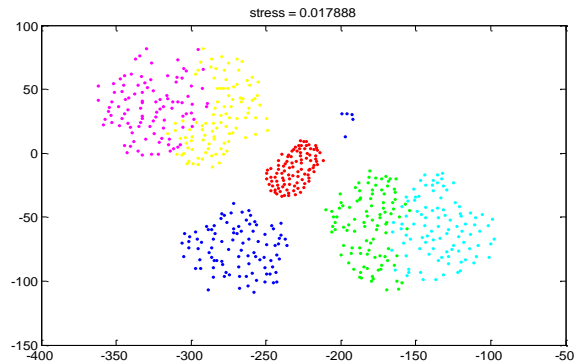

(b)

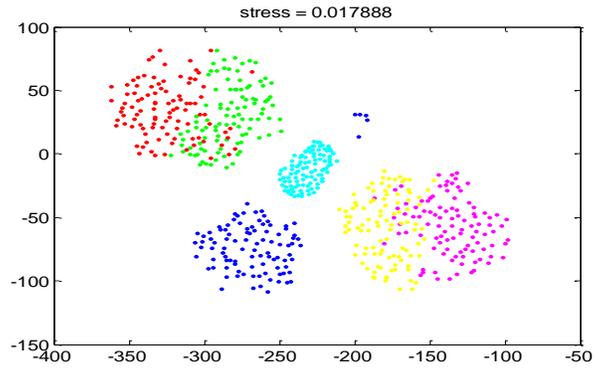

(c)

Figure 5: a. chart dataset projected in two dimensions. Six different colours represent six ground truth clusters. b. Consensus clustering due to CSPA CSPA on a subset selected by CAS selection procedure. c. Consensus clustering due to CSPA on a subset selected by ESDF.

Table 3: Results for Yeast and Ecoli datasets

|  | Ecoli | | | | Yeast | | |
|---|---|---|---|---|---|---|---|
|  | Ensemble1 | Ensemble2 | Ensemble3 | | Ensemble1 | Ensemble2 | Ensemble3 |
| CSPA on all | 0.2105 | 0.2957 | 0.2828 | | 0.0973 | 0.1009 | 0.087 |
| CAS+ CSPA | 0.2562 | 0.3285 | 0.3100 | | 0.1019 | 0.1275 | 0.1112 |
| ESDF+ CSPA | 0.2610 | 0.3367 | 0.3119 | | 0.1129 | 0.1288 | 0.1128 |
| HGPA on all | 0.3453 | 0.3312 | 0.3115 | | 0.1012 | 0.1152 | 0.098 |
| CAS+ HGPA | 0.3622 | 0.3520 | 0.3326 | | 0.1219 | 0.1225 | 0.11 |
| ESDF+ HGPA | 0.3672 | 0.3592 | 0.3337 | | 0.1223 | 0.1243 | 0.12 |

Table 3 shows that both on Ecoli and Yeast datasets CSPA and HGPA produce better consensus when ESDF is used as the ensemble selection procedure rather than CAS.

Table 4: Table for Iris and Vehicle datasets

|  | Iris | | | | Vehicle | | |
|---|---|---|---|---|---|---|---|
|  | Ensemble1 | Ensemble2 | Ensemble3 | | Ensemble1 | Ensemble2 | Ensemble3 |
| CSPA on all | 0.7140 | 0.7140 | 0.7140 | | 0.1207 | 0.1320 | 0.115 |
| CAS+ CSPA | 0.7415 | 0.7415 | 0.7415 | | 0.1452 | 0.1520 | 0.1212 |
| ESDF+ CSPA | 0.8015 | 0.8015 | 0.8015 | | 0.1345 | 0.1399 | 0.116 |
| HGPA on all | 0.6150 | 0.6150 | 0.6150 | | 0.0987 | 0.1009 | 0.099 |
| CAS+ HGPA | 0.7220 | 0.7220 | 0.7220 | | 0.1132 | 0.1234 | 0.101 |
| ESDF+ HGPA | 0.7458 | 0.7458 | 0.7458 | | 0.1201 | 0.1130 | 0.099 |

Table 4 shows that on Iris data both CSPA and HGPA produce better consensus when ESDF is used as the ensemble selection procedure rather than CAS. Whereas, on Vehicle data both CSPA and HGPA produce better consensus when CAS is used as the ensemble selection procedure rather than ESDF.

## VIII. CONCLUSION

In this paper, we demonstrate that it is necessary to select a subset of clustering that can yield better resulting cluster than combine all the available clustering. In our approach we also suggested that clustering combination of diversity and frequency could be a good measure for selecting significant partitions that may contribute to a meaningful consensus. Evaluations on different natural data have shown this approach to be effective in improving clustering accuracy, particularly when true number of clusters of ground truth partition is considered. We show that the proposed method performs well compared to other known ensemble selection methods, although more experiments are needed to assess the real value of the method.

## IX. FUTURE WORK

In this paper we used adjusted rand index as the diversity measure, whereas the method we developed here is independent of diversity measures. Our one part of future work will be to test our method using other diversity measures [8], [1], [15], [5]. Another future direction is to find the optimum size of the ensemble chosen by ESDF for which it gives best result.

## X. ACKNOWLEDGMENT